%% file: main.tex
\documentclass[10pt,twocolumn,letterpaper]{article}

\usepackage{iccv}
\usepackage{times}
\usepackage{epsfig}
\usepackage{graphicx}
\usepackage{grffile} %
\usepackage{amsmath}
\usepackage{amssymb}
\usepackage{booktabs, makecell, tabularx}
\usepackage{subfig}
\usepackage[numbers,sort]{natbib}

\usepackage{rotating}
\usepackage[export]{adjustbox}
\newcommand{\update}[1]{#1}
\newcommand{\change}[1]{#1}

\usepackage{caption}
\usepackage[pagebackref=true,breaklinks=true,letterpaper=true,colorlinks,bookmarks=false]{hyperref}

\input{notations}

\iccvfinalcopy %

\ificcvfinal\fi

\begin{document}

\title{SNARF: Differentiable Forward Skinning for\\ Animating Non-Rigid Neural Implicit Shapes}

\author{Xu Chen$^{1,3}$ \quad Yufeng Zheng$^{1,3}$ \quad Michael J. Black$^{3}$ \quad Otmar Hilliges$^{1}$ \quad Andreas Geiger$^{2,3}$\vspace{0.1cm} \\
 $^1$ETH Z{\"u}rich, Department of Computer Science \quad 
 $^2$University of T{\"u}bingen \\
 $^3$Max Planck Institute for Intelligent Systems, T{\"u}bingen \\
}

\maketitle

\ificcvfinal\thispagestyle{empty}\fi

\begin{abstract}

\input{sections/0_abstract}
\end{abstract}

\input{sections/1_introduction}
\input{sections/2_related_work}

\input{sections/3_method}

\input{sections/4_experiments}
\input{sections/5_conclusion}
\input{sections/6_acknowledgement}

{\small
\bibliographystyle{ieee_fullname}
\bibliography{bibliography_long,bibliography,bibliography_custom}
}

\end{document}

%% file: notations.tex
\newcommand{\pardev}[2]{\frac{\partial {#1}}{\partial {#2}}}

\newcommand{\netparam}{\sigma}
\newcommand{\point}{\mathbf{x}}

\newcommand{\function}[1]{{#1}_{\netparam_{#1}}}
\newcommand{\mfunction}[1]{\mathbf{#1}_{\netparam_{#1}}}

\newcommand{\occ}{o}
\newcommand{\loss}{\mathcal{L}}

\newcommand{\bone}{\boldsymbol{B}}

\newcommand{\jac}{\mathbf{J}}

\newcommand{\figref}[1]{Fig.~\ref{#1}}
\newcommand{\eqnref}[1]{Eq.~(\ref{#1})}
\newcommand{\tabref}[1]{Tab.~\ref{#1}}

\newcommand{\boldparagraph}[1]{\vspace{0.2cm}\noindent{\bf #1:}}

%% file: sections/0_abstract.tex
Neural implicit surface representations have emerged as a promising paradigm to capture 3D shapes in a continuous and resolution-independent manner. However, adapting them to articulated shapes is non-trivial. Existing approaches learn a backward warp field that maps deformed to canonical points. However, this is problematic since the backward warp field is pose dependent and thus requires large amounts of data to learn. To address this, we introduce SNARF, which combines the advantages of linear blend skinning (LBS) for polygonal meshes with those of neural implicit surfaces by learning a forward deformation field without direct supervision. This deformation field is defined in canonical, pose-independent, space, enabling generalization to unseen poses. Learning the deformation field from posed meshes alone is challenging since the correspondences of deformed points are defined implicitly and may not be unique under changes of topology. We propose a forward skinning model that finds all canonical correspondences of any deformed point using iterative root finding. We derive analytical gradients via implicit differentiation, enabling end-to-end training from 3D meshes with bone transformations. Compared to state-of-the-art neural implicit representations, our approach generalizes better to unseen poses while preserving accuracy. We demonstrate our method in challenging scenarios on (clothed) 3D humans in diverse and unseen poses.

%% file: sections/1_introduction.tex
\input{figures/teaser}
\vspace{-1.6em}
\section{Introduction}
Modeling the shape and deformation of articulated 3D objects
has traditionally been achieved by deforming a polygonal mesh via linear blend skinning (LBS) with pose-correctives.
However, meshes are inherently limited by their resolution-to-memory ratio and their fixed topology. Therefore, neural implicit surface representations~\cite{Chen2019CVPR, Mescheder2019CVPR, Michalkiewicz2019ARXIV, Park2019CVPR} have recently attracted much attention because they provide a \update{resolution-independent}, smooth and continuous alternative to discrete meshes. However, updating \update{an} implicit surface representation as a function of \update{the} underlying pose changes is challenging \change{since} it requires modifying a continuous function rather than a discrete set of points. 

To address this, we propose \textit{SNARF (Skinned Neural Articulated Representations with Forward skinning)}, a novel approach to learning articulated 3D shapes represented by neural implicit surfaces directly from 3D watertight meshes and corresponding bone transformations with no need for supervision via pre-defined skinning weights.
SNARF
combines the simplicity of skeletal-driven deformation of LBS with the fidelity and topological flexibility of implicit surfaces, enabling animation of complex human bodies as shown in \figref{fig:teaser}.
Moreover, SNARF goes beyond LBS by conditioning the neural shape on poses to capture pose-dependent \update{non-linear} deformations.
\update{The main challenge is to express the mapping between surface points in canonical pose and their deformed counterparts.}
Existing approaches attempt to learn shape in the canonical pose and a \textit{backward} deformation field, transforming deformed points to the canonical pose~\cite{Deng2020ECCV,Niemeyer2019ICCV,Park2020ARXIV,Pumarola2020ARXIV}.
However, as illustrated in \figref{fig:backward_vs_forward}, backward skinning is problematic \change{since} the deformation field depends on the pose of the deformed object, %
limiting generalization to unseen poses.
\input{figures/backward_vs_forward}

To tackle this problem, 
we devise a method that learns a dense \emph{forward} skinning weight field without requiring direct supervision. 
Once learned, this skinning field can be leveraged to generate shape deformations even for poses outside of the training set.
However, to jointly learn the forward skinning field and the object shape from posed meshes alone, we must establish the correspondence of any 3D point in deformed space to the undeformed space. 
Yet, this requires the availability of the \textit{backward} mapping which is only implicitly defined and has no analytical solution. 

To overcome this issue, we propose a forward skinning model that exploits an iterative root finding algorithm to find the corresponding canonical point for any deformed point. Our approach is able to retrieve multiple correspondences for any deformed point and therefore naturally handles topology changes.
We further derive the gradients of our forward skinning module, hence making it differentiable and enabling end-to-end learning of the canonical shape and skinning weights jointly from deformed observations. 
Importantly, and in contrast to prior work, our method does not require any a priori skinning weights or pose correctives defined on the surface and hence can be applied in scenarios where pre-rigged mesh models are not available.
We experimentally demonstrate that our method is able to generate \update{high-quality} shapes with arbitrary desired bone transformations, even those far beyond the training distribution, where other recent methods like NASA~\cite{Deng2020ECCV} fail. Since our approach operates in continuous space, it enables reconstruction of fine geometric details. 
By conditioning the neural implicit function on poses, our method faithfully models local pose-dependent deformations, \eg, the movement of clothing or soft tissue. 
\update{Our code is available at \url{github.com/xuchen-ethz/snarf}.}

%% file: figures/teaser.tex
\begin{figure}
    \centering
    \includegraphics[width=\linewidth,trim=0 5 0 5, clip]{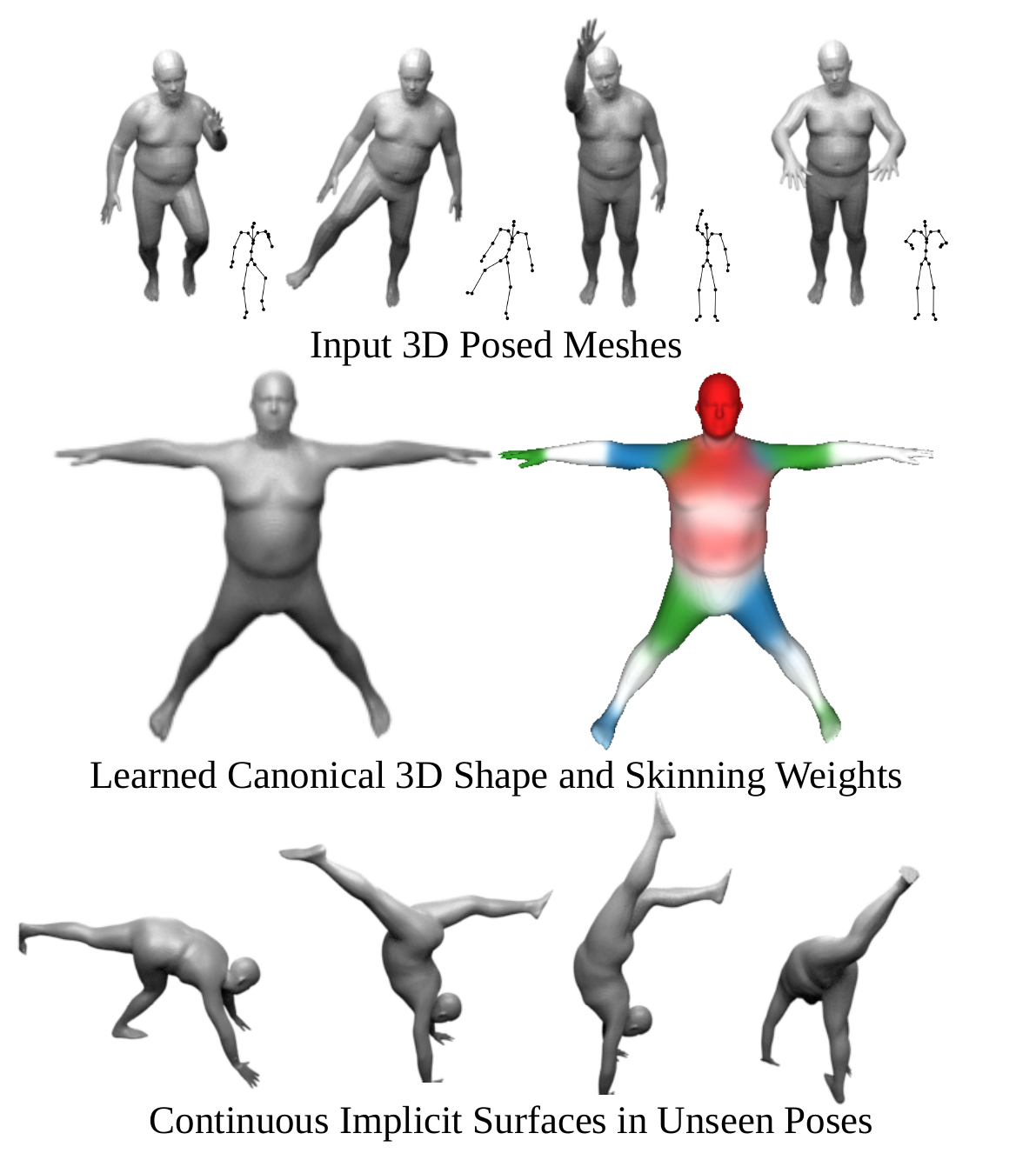}
    \caption{\textbf{SNARF:} 
    From a sequence of posed meshes (top),
    we learn a neural implicit 3D shape and a skinning field in canonical pose (middle) without supervision of skinning weights or part correspondences.
    Learned forward skinning enables generalization to unseen poses (bottom) %
    while capturing local details via pose conditioning.
    \vspace{-1.4em}
    }
    \label{fig:teaser}
\end{figure}

%% file: figures/backward_vs_forward.tex
\begin{figure}
    \centering
    \includegraphics[width=\linewidth,trim=0 4 0 0, clip]{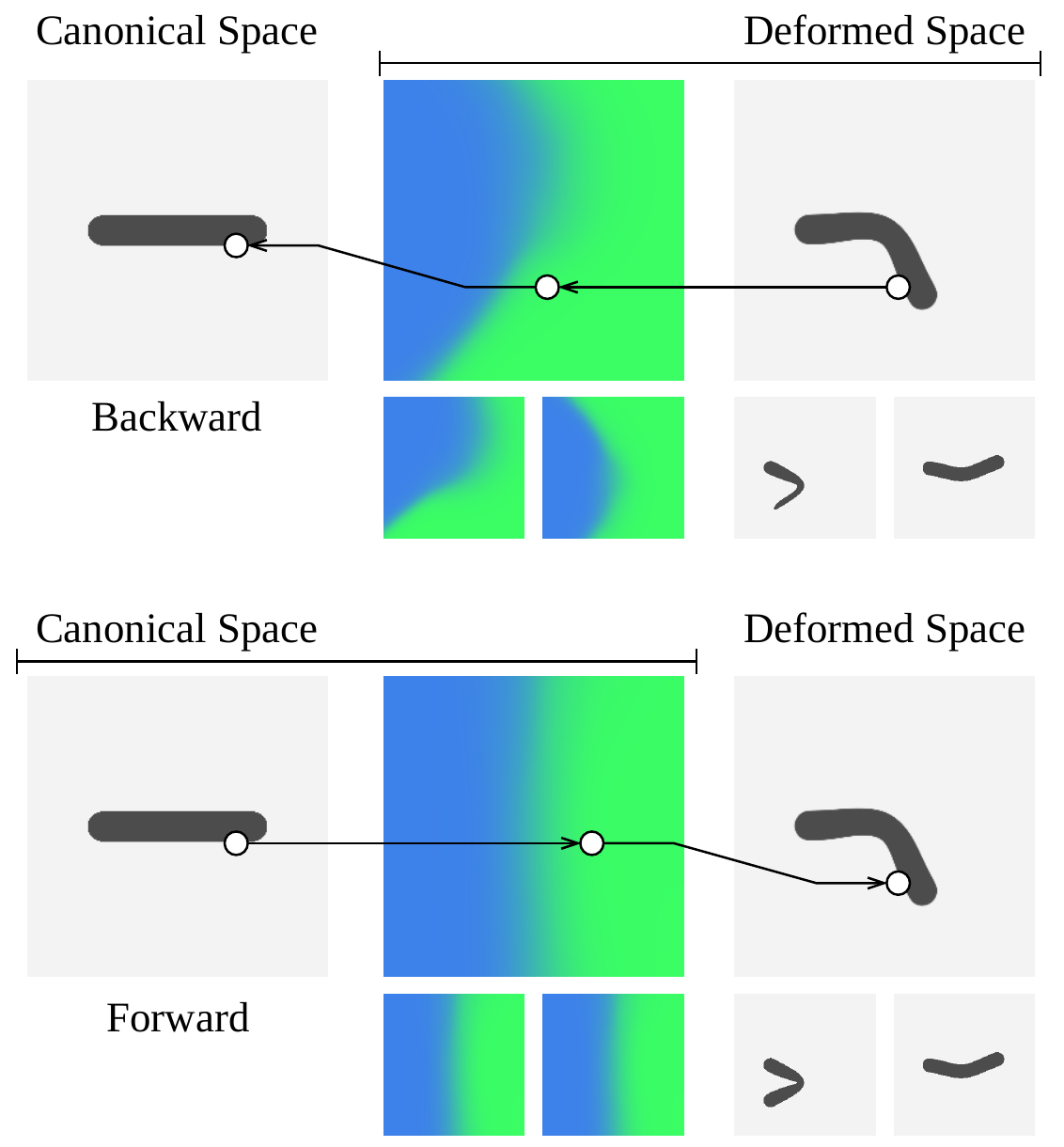}
    \caption{\textbf{Forward vs.~Backward Skinning.}
\change{Forward skinning weights are defined in pose-independent canonical space and therefore naturally generalize to unseen poses as the one in the bottom left panel. In contrast, backward skinning weights are defined in pose-dependent deformed space and thus struggle with unseen poses.} \vspace{-1.4em}}
    \label{fig:backward_vs_forward}
\end{figure}

%% file: sections/2_related_work.tex
\section{Related Work}
\vspace{-0.5em}
\boldparagraph{Skinning Polygonal Meshes}
Modeling the deformation of non-rigid and articulated 3D objects is a fundamental problem in computer vision and graphics with many applications.
Traditionally, this problem is formulated for polygonal meshes and is referred to as skinning. Skinning enables deformation of a high-resolution surface mesh with low-order control primitives such as skeletal bones. The most common approach is linear blend skinning (LBS), which models each mesh vertex's deformation as a convex combination of input bone transformations as defined by skinning weights.
These skinning weights are typically defined by an artist or learned from data.
LBS produces \update{well-known} artifacts that many methods attempt to address, \eg with dual quaternion blend skinning~\cite{Kavan2008SIGGRAPH} or multi-weight enveloping~\cite{Merry2006TOG,Wang2002SIGGRAPH}.
The key concept is to define pose-dependent ``corrective blend shapes" that are added to a shape such that, when it is posed, the LBS errors are minimized \cite{Lewis:2000:PSD,rouet1999method}.
Classically, these ``pose correctives" are artist defined, though they can also be learned \cite{Loper2015SIGGRAPH}.
Here we extend the concept of LBS and pose correctives to neural implicit surface representations.

Learning both blend weights and rigs from examples has a long history, starting with James and Twigg \cite{James:1995}. 
Specifically for human bodies, numerous learning methods have been proposed, many of which learn the LBS weights \cite{Hasler:2010,Loper2015SIGGRAPH,Osman:STAR:2020,Xu:GHUM:2020}.
Recent methods attempt to disentangle shape and pose in an unsupervised fashion given registered training meshes \cite{Jaing:TVCG:2020,Zhou:ECCV:2020}.
RigNet \cite{Zu:RigNet:2020} uses a deep network to learn both articulated rigs and skinning weights jointly.
NeuroSkinning \cite{Liu:ToG:2019} also uses a deep network to learn blend weights and can cope with complex surface topology. 
In contrast to us, these methods require a large dataset of rigged models with hand-painted skinning weights and do not consider implicit surface representations.

\boldparagraph{Neural Implicit Shapes}
Neural implicit shape representations can model complex shapes with arbitrary topology in a continuous fashion.
Given a 3D location, these networks regress the distance to the surface~\cite{Park2019CVPR}, occupancy probability~\cite{Mescheder2019CVPR}, color~\cite{Oechsle2019ICCV} or radiance~\cite{Mildenhall2020ECCV} of a 3D point. 
Conditioning on local information such as 2D image features or 3D point cloud features has been shown to yield more detailed reconstructions~\cite{Chibane2020CVPR, He2020NIPS, Peng2020ECCV, Saito2019ICCV, Saito2020CVPR}. 
While early methods require watertight meshes for training, several recent approaches have demonstrated unsupervised training from raw 3D points clouds~\cite{Atzmon2020CVPR, Gropp2020ICML, Saito2021CVPR} or images~\cite{Mildenhall2020ECCV, Niemeyer2020CVPR, Sitzmann2019NIPS, Yariv2020NIPS}. \update{A current limitation of most existing implicit models is that they do not support high-quality skeletal deformation.
Our method addresses this key limitation, enabling learning and generation of realistic skeletal deformations of neural implicit surfaces.}
\boldparagraph{Deformable Neural \update{Shapes}}
\update{
Compared to
meshes, deforming neural implicit shapes is more challenging as one needs to deform continuous space rather than a fixed set of vertex points. Very recently, various approaches have been proposed to model backward deformation fields~\cite{Jeruzalski2020ARXIV, Niemeyer2019ICCV,Park2020ARXIV,Pumarola2020ARXIV}.}
These fields map points in deformed space to canonical ones, where geometric properties (e.g.~occupancy) are queried from a canonical shape network.
The deformation field is modeled as a neural network that outputs velocity~\cite{Niemeyer2019ICCV}, translation~\cite{Pumarola2020ARXIV} or rigid transformation~\cite{Park2020ARXIV} and is jointly trained with the canonical occupancy network using observations in deformed space.
NiLBS~\cite{Jeruzalski2020ARXIV} learns skinning weights for each point and then derives the deformation via LBS according to the bone transformations.
An inherent limitation of learned backward deformation, however, is poor generalization to unseen poses. 
As illustrated in Fig.~\ref{fig:backward_vs_forward}, backward deformation fields are defined in deformed space and, hence, inherently deform with pose. Thus, the network must memorize deformation fields for different spatial configurations, making it difficult to generate deformations that have not been seen during training.

\boldparagraph{Part-based Models}
In recent work, NASA~\cite{Deng2020ECCV}, proposes to represent a 3D human body model as a combination of independent parts, each of which is represented by an occupancy network \cite{Mescheder2019CVPR}.
Rigidly transforming these parts according to the input bone transformations produces deformed shapes.
While such a formulation preserves the global structure after articulation, the continuity of surface deformations is violated, causing artifacts at intersections of body parts.
Although each part can learn to deform itself to partially compensate for this undesired effect, noticeable artifacts remain, particularly for poses that are beyond the training distribution.
\update{Moreover, NASA requires ground-truth surface skinning weights to learn correct part assignments.
In contrast to NASA, our method learns forward skinning weights without such supervision and captures pose-dependent deformations.}

\update{More generally, \change{the previous approaches} suffer from artifacts due to overly simple assumptions about deformation or do not generalize well to unseen poses \change{as shown in \figref{fig:toy}}. In contrast, SNARF generates continuous shapes in arbitrary poses, even those far beyond the training distribution, by learning pose-independent forward skinning weights and pose-dependent correctives in canonical space.}

\input{figures/pipeline}

\boldparagraph{3D Human Avatars}
While more general, we demonstrate our approach on the problem of learning and animating realistic 3D human avatars.
\update{
Recent~\cite{Alldieck20183DV,Alldieck2019ICCV,Zheng2019ICCV,Anguelov:ToG:2005, Bhatnagar2020NIPS} and concurrent works~\cite{Peng2021CVPR, Weng2020ARXIV,Raj2020ARXIV,Shysheya2019CVPR, Mihajlovic2021CVPR, Ma2021CVPR, Wang2021CVPR, Habermann2021SIGGRAPH, Burov2021ICCV, Peng2021ICCV, Ma2021ICCV, Liu2021ARXIV, Saito2021CVPR, Wang2021ARXIV} on learning 3D human models typically require a template mesh model with fixed topology, e.g. SMPL~\cite{Loper2015SIGGRAPH}, or are limited in resolution due to the underlying 3D representation~\cite{Alldieck20183DV,Zheng2019ICCV,Alldieck2019ICCV,Weng2020ARXIV}.
}
In contrast, our method is able to represent articulated shapes at high fidelity without strong prior assumptions about the object's shape. This allows us to better model deformations of objects with more flexible topology, \eg, humans in clothing.

%% file: figures/pipeline.tex
\begin{figure*}
    \centering
    \includegraphics[width=\textwidth]{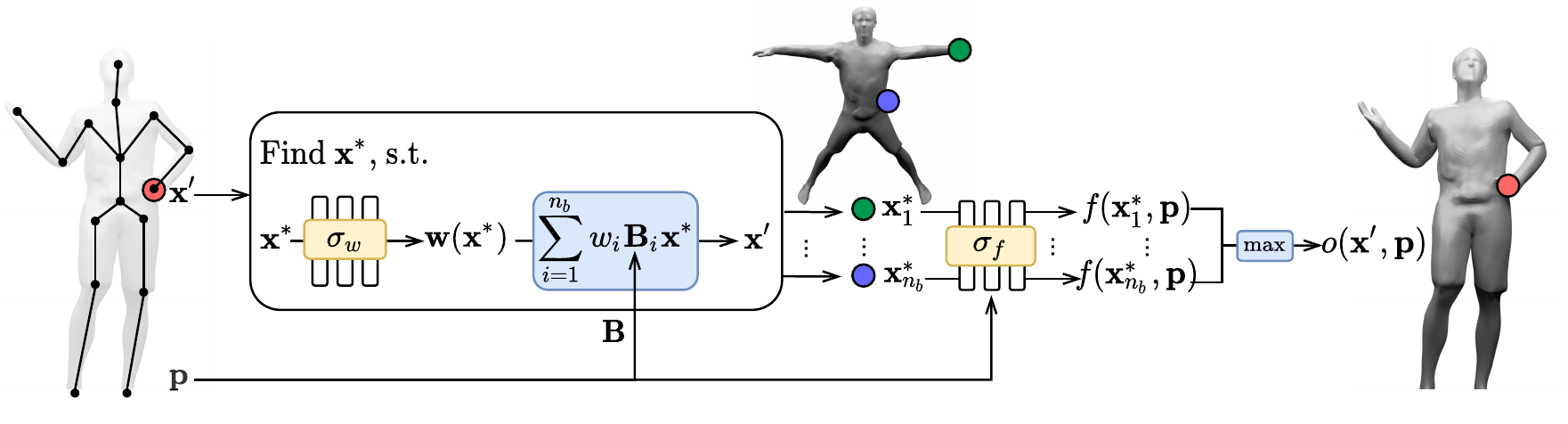}
    \vspace{-2.5em}
    \caption{\textbf{Generating Deformed Shapes with Forward Skinning.} Given a query point in deformed space $\point'$, our method first finds its canonical correspondences $\point^*$ which satisfy the forward skinning equation \eqref{equ:lbs} via iterative root finding. Multiple correspondences may exist due to topological changes, which can be reliably found by initializing the root finding algorithm with multiple starting points derived from the bone transformations. The canonical occupancy network $\function{f}$ then predicts the occupancy probabilities at $\{\point^*\}$ which are finally aggregated to yield the occupancy probability of the query point $\point'$.\vspace{-1em}}
    \label{fig:pipeline}
\end{figure*}

%% file: sections/3_method.tex
\section{Method}

In this section, we first define our representation for the canonical shape and forward skinning weights. Next, we introduce our forward mapping and derive the gradients for learning the canonical shape representation and skinning weights in an end-to-end manner. 

\subsection{Representation}
We represent an articulated object by its shape and skinning weights in canonical space.
\update{Similar to classical approaches like SMPL, we split the problem into LBS with pose-independent skinning weights and pose-dependent non-linear deformations.}
LBS captures many important aspects of the shape change, thus the pose-dependent model only has to learn a corrective.
This makes training with limited data feasible and aids generalization to unseen poses.

\boldparagraph{Shape} We use a neural network to predict the occupancy probability for any input 3D point $\point$ in canonical space. To model pose-dependent local deformations such as wrinkles or soft tissue, we inject the object pose $\mathbf{p}$ as additional input:
\begin{align}
\function{f}: \mathbb{R}^3 \times \mathbb{R}^{n_{p}} \rightarrow [0,1] .
\end{align}
Here, $\sigma_f$ are the network parameters and $n_p$ is the dimensionality of the pose condition $\mathbf{p}\in\mathbb{R}^{n_p}$ which we specify in terms of joint angles.
The canonical shape is implicitly defined as the $0.5$ level set of the neural function $\mathcal{S}$:
\begin{align}
\mathcal{S} = \{ \point \mid \function{f}(\point,\mathbf{p}) = 0.5\} .
\end{align}

\boldparagraph{Neural Blend Skinning} 
We model the non-rigid deformation induced by skeleton changes using linear blend skinning (LBS). Towards this goal, we represent an LBS weight field in canonical space using a second neural network:
\begin{align} 
\mfunction{w}: \mathbb{R}^3 \rightarrow \mathbb{R}^{n_b} ,
\end{align}
where $\sigma_w$ are the network parameters and $n_b$ denotes the number of bones. Following traditional LBS, we enforce the weights $\mathbf{w}=\{w_1,\dots,w_{n_b}\}$ of each point $\point$ to satisfy $w_i \geq 0$ and $\sum_{i} w_i = 1$ using a softmax activation function. Note that $\mfunction{w}$ does not depend on the pose $\mathbf{p}$.

Given the LBS weights $\mathbf{w}$ of a 3D point $\point$ and the bone transformations $\bone=\{\bone_1,\dots,\bone_{n_b}\}$ corresponding to a particular body pose $\mathbf{p}$, the deformed point $\point'$ is determined by the following convex combination:
\begin{align}
\point' = \mathbf{d}_{\sigma_w}(\point,\bone) = \sum_{i=1}^{n_{\text{b}}}{w}_{\sigma_{w},i}(\point) \cdot \bone_i\cdot \point
. \label{equ:lbs}
\end{align}

\update{Note that the canonical pose is a free hyper-parameter. Empirically, we found the canonical pose shown in \figref{fig:teaser} to work well and used it for all experiments on human shapes.}

\subsection{Differentiable Forward Skinning}
To predict the occupancy probability $\occ_\point'$ of an observed 3D point $\point'$ in deformed space, we must first determine the canonical correspondence $\point^*$ of the deformed query $\point'$ in order to evaluate the occupancy $\occ(\point',\mathbf{p}) = f(\point^*,\mathbf{p})$ with the canonical occupancy network.

At the core of our forward skinning approach lies the problem of finding canonical correspondence $\point^*$ of any query point $\point'$. This is non-trivial \change{because} (i) their relationship is defined implicitly via \eqnref{equ:lbs} without an analytical inverse form, and (ii) multiple canonical points might correspond to the same deformed point as space can overlap after warping (cf.~\figref{fig:pipeline}). To address this problem, we propose a procedure that is able to retrieve all potential canonical correspondences $\{\point_i^*\}$ of any deformed point $\point'$ from the implicitly defined relationship and then composite these correspondences using standard operations for implicit shape composition. An overview is provided in \figref{fig:pipeline}.

\boldparagraph{Correspondence Search} 
Unlike backward skinning, forward skinning defines the canonical correspondence $\point^*$ of $\point'$ implicitly as the root of the following equation
\begin{align}
\mathbf{d}_{\sigma_w}(\point,\bone) - \point' = \mathbf{0},
\label{equ:root_equ}
\end{align}
which cannot be solved in closed form. The solution of \eqnref{equ:lbs} can be attained numerically via standard Newton or quasi-Netwon methods:
\begin{align}
\point^{k+1} &= \point^{k} -(\jac^k)^{-1}\cdot ( \mathbf{d}_{\sigma_w}(\point^k, \bone) - \point' ) ,
\label{equ:broyden}
\end{align}
where $\jac$ is the Jacobian matrix of $\mathbf{d}_{\sigma_w}(\point^k, \bone) - \point'$. To prevent computing the Jacobian at each iteration, we apply Broyden's method \cite{Broyden1965BOOK} using a low-rank approximation of $\jac$.

\boldparagraph{Handling Multiple Correspondences}
We find multiple roots $\{\point_i^*\}$ by initializing the optimization procedure with different starting locations and exploiting the local convergence of iterative root finding. The initial states $\{\point_i^0\}$ are thereby obtained by transforming the deformed point $\point'$ rigidly to the canonical space for each of the $n_{b}$ bones, \update{and the initial Jacobian matrices $\{\jac_i^0\}$ are the spatial gradients of the LBS weight field at \change{the} corresponding initial states:}
\begin{align}
\update{
\point^{0}_i = \bone_i^{-1} \cdot \point' \quad
J^{0}_i = \pardev{\mathbf{d}_{\sigma_w}(\point, \bone)}{\point}\bigg|_{\point = \point_i^0} }
\label{equ:init}
\end{align}
Initial states that are far from the optima lead to either convergence to one of the optima and can be safely included for further computation, or divergence, and can therefore be easily discarded by thresholding. Consequently, we define the final set of correspondences as:
\begin{align}
\mathcal{X}^* = \left\{\point^*_i \mid \left \| \mathbf{d}_{\sigma_w}(\point^*_i, \bone) - \point' \right \|_2 < \epsilon \right\} ,
\label{equ:root_set}
\end{align}
where $\epsilon$ is the convergence threshold which we set to $10^{-5}$ in our experiments. This allows us to retrieve all canonical correspondences of any deformed point $\point'$ even under topological changes which induce one-to-many mappings. 

Note that if any of the canonical correspondences is occupied, the deformed point $\point'$ is occupied as well. Thus, the maximum over the occupancy probabilities of all canonical correspondences gives the final occupancy prediction:
\begin{align}
\occ(\point',\mathbf{p}) = \max_{\point^* \in \mathcal{X}^*} \{  \function{f}(\point^*,\mathbf{p}) \} .
\label{equ:compose}
\end{align}
This union operator is commonly used to composite independent shapes~\cite{Ricci1973BOOK}. Similar to NASA~\cite{Deng2020ECCV}, in practice we use softmax instead of a hard maximum to allow gradients to back-propagate to all canonical correspondences.

\subsection{Training Losses}
\update{
Our model is trained \change{via} minimizing the binary cross entropy loss $\loss_{BCE}(\occ(\point',\mathbf{p}), \occ_{gt}(\point'))$ between the predicted occupancy of the deformed points $\occ(\point',\mathbf{p})$ and the corresponding ground-truth $\occ_{gt}(\point')$ for all posed 3D meshes of a single subject.
\change{In addition, we apply two auxiliary losses during the first epoch to bootstrap training}. We randomly sample points along the bones \change{that connect} joints in canonical space and encourage their occupancy probabilities to be one. Moreover, we encourage the skinning weights of all joints to be equal to $0.5$ for their respective two neighboring bones. No ground truth skinning weights or part segmentations are required by our method.}

\subsection{Gradients}

During training, we must determine the gradient of the overall loss $\mathcal{L}$ \wrt the network parameters $\sigma = \{\sigma_f,\sigma_w\}$. For the occupancy network $\function{f}$, the gradient is given by
\begin{align}
\pardev{\loss}{\sigma_f} = \pardev{\loss}{\occ} \cdot \pardev{\occ}{\function{f}} \cdot \pardev{\function{f}}{\sigma_f}
\end{align}
which can be easily obtained by backpropagating gradients through the corresponding computation graph.
For the LBS weight field $\mfunction{w}$, the gradient is given by
\begin{align}
\pardev{\loss}{\sigma_w} = \pardev{\loss}{\occ} \cdot \pardev{\occ}{\function{f}} \cdot \pardev{\function{f}(\point^*)}{\point^*} \cdot \pardev{\point^*}{\sigma_w}
\end{align}
where $\point^*$ is the root as defined in \eqnref{equ:root_set} and %
the last term can be analytically obtained via implicit differentiation:
\begin{align}
& \mathbf{d}_{\sigma_w}(\point^*,\bone) - \point' = \mathbf{0}  \\
\Leftrightarrow ~& \pardev{\mathbf{d}_{\sigma_w}(\point^*,\bone)}{\sigma_w} + \pardev{\mathbf{d}_{\sigma_w}(\point^*,\bone)}{\point^*} \cdot \pardev{\point^*}{\sigma_w} = \mathbf{0}\\
\Leftrightarrow ~& \pardev{\point^*}{\sigma_w} = - \left(\pardev{\mathbf{d}_{\sigma_w}(\point^*,\bone)}{\point^*} \right)^{-1}\cdot \pardev{ \mathbf{d}_{\sigma_w}(\point^*,\bone)}{\sigma_w} . 
\label{equ:def_grad}
\end{align}

%% file: sections/4_experiments.tex
\section{Experiments}
\label{sec:exp}

We first conduct toy experiments on synthetic 2D data to analyze different methods and model design choices in a controlled setting. Next, we apply our approach to model minimally clothed human bodies and compare it to 
NASA~\cite{Deng2020ECCV} and other self-implemented baselines. Finally, we demonstrate that our method can handle clothed humans, generalizing well to unseen poses.

\subsection{Datasets}
\noindent We use the following datasets in our experiments:

\boldparagraph{2D Stick}
We simulate a 2D stick articulated by two bones.
We set the true skinning weights of each point as the
the inverse of its distance to each bone.
To simulate topology changes, we include a further rigid object. While this object is separate in canonical space, the two may intersect in posed space and therefore cause topology changes to simulate human self-contact or object interaction.

\boldparagraph{Minimally Clothed Humans} Following NASA~\cite{Deng2020ECCV}, we use the DFaust~\cite{Bogo2017CVPR} subset of AMASS~\cite{Mahmood2019ICCV} for training and evaluating our model on SMPL meshes of people in minimal clothing. This dataset covers 10 subjects of varying body shapes. For each subject, we use 10 sequences, from which we randomly select one sequence for validation, using the rest for training. For each frame in a sequence, 20K points are sampled, among which, half are sampled uniformly in space and half are sampled in near-surface regions by \change{first} applying Poisson disk sampling on the mesh surface, followed by adding isotropic Gaussian noise with $\sigma=0.01$ to the sampled point locations. Besides the ``within distribution'' evaluation on DFaust, we also include another subset named PosePrior~\cite{Akhter2015CVPR} from AMASS for \change{an} ``out of distribution'' evaluation. This dataset contains natural, more challenging, poses beyond those in DFaust.

\boldparagraph{Clothed Humans} We use the registered meshes from CAPE~\cite{Ma2020CVPR} and corresponding joints and bone transformations derived from the accompanied SMPL model registration to train our model. We use 8 subjects from the dataset with different clothing types including short/long lower body clothing and short/long upper body clothing.
We train a model for each subject and clothing condition.

\subsection{Baselines}
\noindent We consider the following baselines in our evaluation. For \update{``Back-LBS'',``Back-D''} and ``Pose-ONet'' we use the same training losses and hyperparameters as in our approach. 

\boldparagraph{Pose-Conditioned Occupancy Networks (Pose-ONet)}
This baseline extends Occupancy Networks~\cite{Mescheder2019CVPR} by directly concatenating the pose input to the occupancy network.

\boldparagraph{Backward Skinning \update{(Back-LBS)}} 
This baseline implements the concept of backward skinning similar to~\cite{Jeruzalski2020ARXIV}. A network takes a deformed point and pose condition
as input and outputs the skinning weights of the deformed point. The deformed point is then warped back to canonical space via LBS and the canonical correspondence is fed into the canonical shape network to query occupancy.

\update{
\boldparagraph{Backward Displacement (Back-D)} 
This baseline directly predicts the displacement from deformed space to canonical space, similar to D-NeRF~\cite{Jeruzalski2020ARXIV}.}

\boldparagraph{NASA}
NASA~\cite{Deng2020ECCV} models articulated human bodies as a composition of multiple parts, each of which transforms rigidly and deforms according to the pose. Note that in contrast to us, NASA requires ground-truth skinning weights for surface points as supervision. 
We use the official NASA implementation provided by the authors.

\boldparagraph{Piecewise}
For evaluation on the 2D toy dataset, we created a variant of NASA for 2D which we refer to as ``Piecewise''.

\subsection{Results on 2D Stick Dataset}

\input{results/toy/visuals}

\input{results/toy/plot}

\input{results/3d_naked/numbers_with_dnerf}

\input{results/3d_naked/visuals_with_dnerf}

\input{results/3d_clothed/visuals}

\noindent For our results on the simple 2D stick dataset, we do not use local pose-conditioning as the shape does not locally deform with pose. We consider the following three settings:

\boldparagraph{Extrapolation}
An essential requirement for articulated models is the ability to deform into arbitrary poses. 
In this setting, we generate training data using the articulated 2D stick with joint angles from the interval $[-60,60]^\circ$. At test time, the models are tasked to generate deformed shapes with larger joint angles in $[-120, 60]^\circ\cup[60, 120]^\circ$. 
\figref{fig:toy} and \figref{fig:toy_plot} (left) show our results.
While our forward skinning model follows the ground truth closely, Pose-ONet fails to generate a meaningful structure as it learns a direct mapping from poses to shapes and thus cannot produce unseen shapes. By disentangling deformations from shapes, \update{Back-LBS} \change{preserves the structure better}, but the learned pose-dependent skinning weights do not generalize. The piecewise model \update{(Piecewise)} generates the correct global \change{pose configuration} but exhibits visible artifacts as the rigidity assumption is violated at the joint.

\boldparagraph{Topological Changes}
To simulate topological changes, we include a rigid object but otherwise keep the setting the same as in the previous experiment. Changing topology is challenging for \update{Back-LBS} \change{since} it is not able to model one-to-many backward correspondences. To compensate for this, the occupancy field gets distorted as shown in \figref{fig:toy}. In contrast, our model gracefully handles topological changes, as also shown quantitatively in \figref{fig:toy_plot} (left).

\boldparagraph{Interpolation}
To assess interpolation performance, we evaluate the accuracy of the generated shapes with angles sampled continuously from $[-60,60]^\circ$ while increasing the sampling step size of the training poses.
As shown in \figref{fig:toy_plot} (right), with increasing difficulty, the gap between the baseline methods (Pose O-Net and \update{Back-LBS}) and ours becomes larger.
An exception is the piecewise model \update{(Piecewise)}, whose performance is invariant to the training sample density, but instead exhibits artifacts at part intersections.

\subsection{Results on Minimally Clothed Humans}

Following NASA~\cite{Deng2020ECCV}, we now consider the more challenging case of modeling articulated 3D human bodies. Human bodies are challenging due to their complex skeletal structure and local deformations that are non-linearly dependent on the bone transformations.
While NASA requires ground-truth skinning weights as additional supervision, our method does not require such knowledge.

\boldparagraph{Within Distribution}
Overall, all methods perform well in this relatively simple setting, as \update{shown} in \tabref{tab:3d_naked}. However, our method still provides an improvement over all baselines. In particular, compared to  NASA~\cite{Deng2020ECCV}, we improve the IoU of uniformly sampled points by $1.2\%$ and the IoU of near-surface points by $4.6\%$.
This improvement can also be observed in the qualitative results \figref{fig:3d_naked}. Our method produces \update{bodies} with smooth surfaces and correct \update{poses}. In contrast, NASA suffers from discontinuous artifacts near joints. \update{Back-D, Back-LBS} and Pose-ONet suffer from missing body parts.

\boldparagraph{Out of Distribution}
In this setting, we test the trained models on a different dataset, PosePrior~\cite{Akhter2015CVPR}, to assess the performance in more realistic settings, where poses can be far from those in the training set. Similar to the observations in the 2D toy setting, unseen poses may cause drastic performance degradation to the baseline methods as shown in~\tabref{tab:3d_naked}. In contrast, our method degrades gracefully despite test poses being drastically different from training poses and very challenging. Hence, the performance gap on IoU surface between our method and NASA increases from $4.6\%$ to $20.4\%$. As can be seen in~\figref{fig:3d_naked}, our method generates natural shapes for the given poses while NASA fails to generate correctives at bone intersections for unseen poses, leading to noticeable artifacts. \update{Pose-ONet and Back-D fail} to generate meaningful shapes and \update{Back-LBS} produces distorted bodies due to incorrect skinning weights.

\boldparagraph{Learned Skinning Weights}
We demonstrate our learned skinning weights in \figref{fig:teaser}.
Our model learns plausible skinning weights  with smooth transitions for all moving body parts, reflecting the correct body part assignment. More results can be found in the supplementary material.

\input{figures/pose_deformation}

\subsection{Results on Clothed Humans}

Our method can also be applied to modeling clothed humans. We train SNARF using meshes from the CAPE dataset. The results are shown in \figref{fig:3d_clothed}. Our method is able to model different clothing types with flexible topology and generates realistic results in novel poses with plausible local details, such as wrinkles. The clothing deforms naturally with the body pose, except for very extreme poses where prediction quality degrades gracefully. \update{\figref{fig:pose_deformation} shows the canonical shapes corresponding to different body poses.}

%% file: results/toy/visuals.tex
\begin{figure}[!t]
	\centering
	\newcommand{\mywidth}{0.17\linewidth}
    \setlength{\tabcolsep}{1pt}

	\begin{tabular}{ccccccc}
	    &&
	    {\small Pose-ONet} &
	    {\small Piecewise} &
	    {\small \update{Back-LBS}} &
	    {\small Ours} &
	    {\small GT}
        \\

	    &\rotatebox{90} {\change{2 bones}}
		&\includegraphics[width=\mywidth]{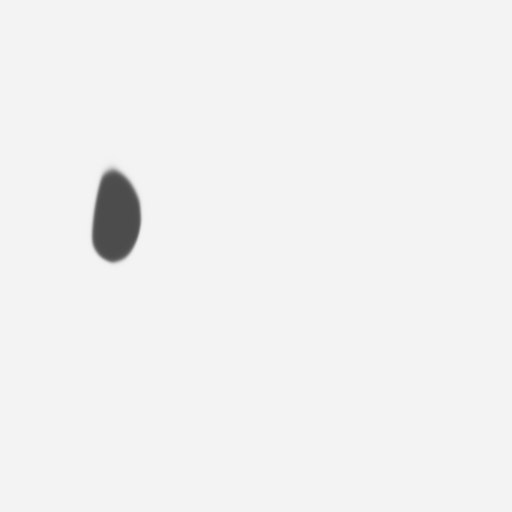} &
		\includegraphics[width=\mywidth]{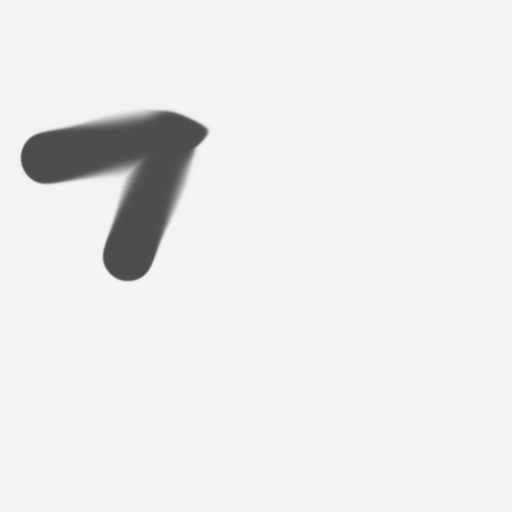} &
		\includegraphics[width=\mywidth]{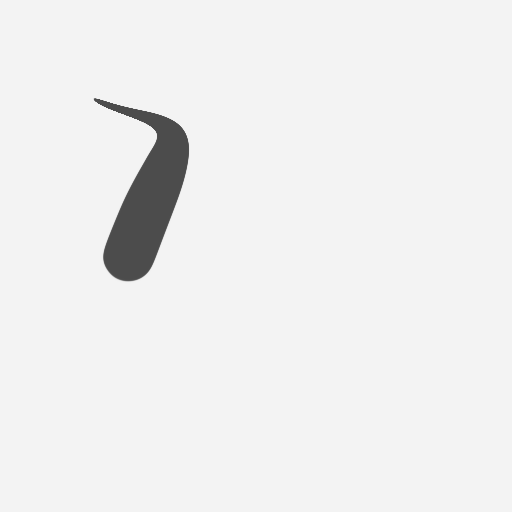} & 
   		\includegraphics[width=\mywidth]{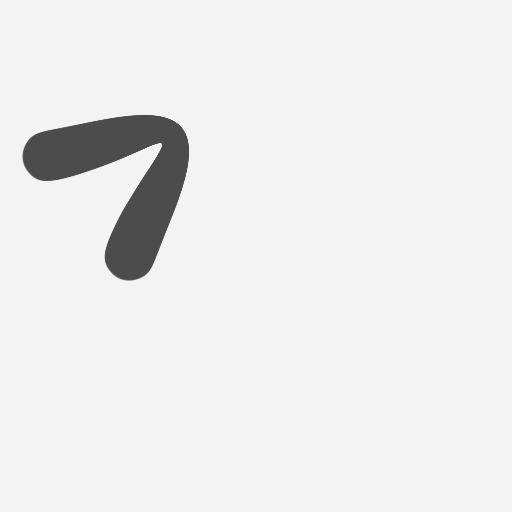} &
		\includegraphics[width=\mywidth]{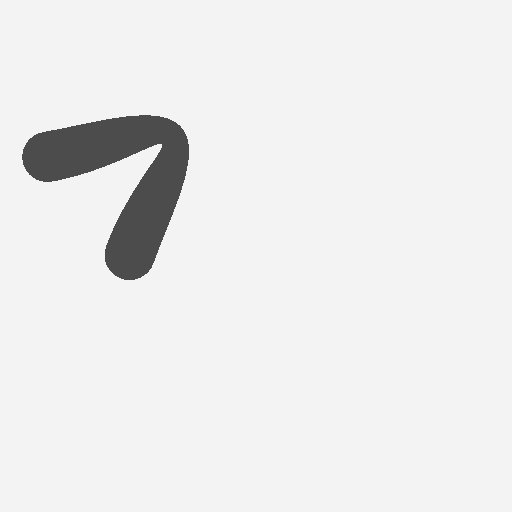}
		\\[-0.15em]

	   \rotatebox{90} {\change{2 bones}} & \rotatebox{90} {\change{1 object}}
   		&\includegraphics[width=\mywidth]{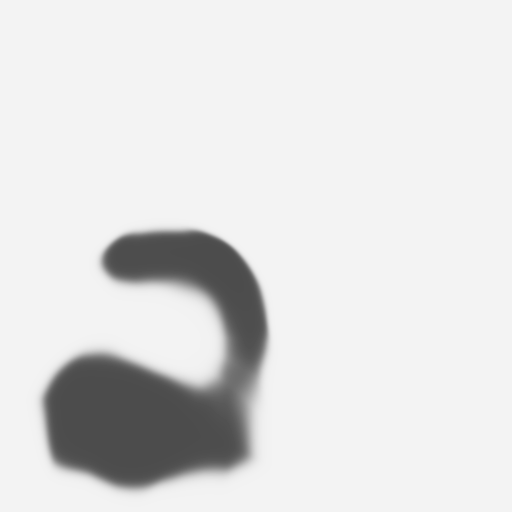} & 
		\includegraphics[width=\mywidth]{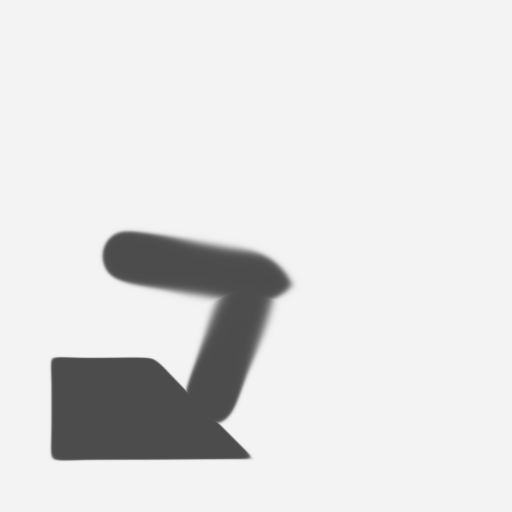} &
		\includegraphics[width=\mywidth]{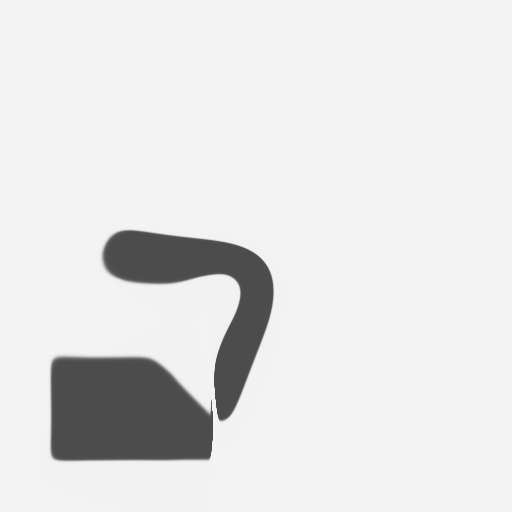} & 
		\includegraphics[width=\mywidth]{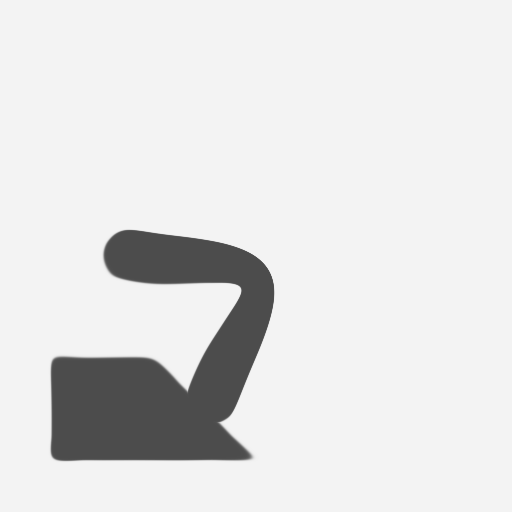} &
		\includegraphics[width=\mywidth]{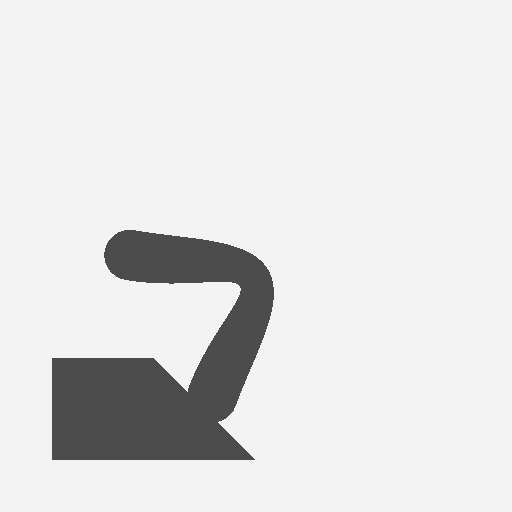}
		\\

	\end{tabular}
	\caption{
		\textbf{Qualitative Results on 2D Toy Experiment.} \emph{Row 1 (2 bones):} Our deformed shape appears similar to the ground-truth. In contrast, \update{Back-LBS} and Pose-ONet produce distorted shapes. The \update{piecewise} rigid model \update{(Piecewise)} leads to artifacts around bone intersections. \emph{Row 2 (2 bones + 1 rigid object):} Our forward skinning algorithm can handle topology changes while artifacts at the intersection are noticeable in the result of the backward skinning baseline \update{(Back-LBS)}.\vspace{-1em}
	}
	\label{fig:toy}
\end{figure}

%% file: results/toy/plot.tex
\begin{figure}[!t]
    \centering
    \includegraphics[width=\linewidth]{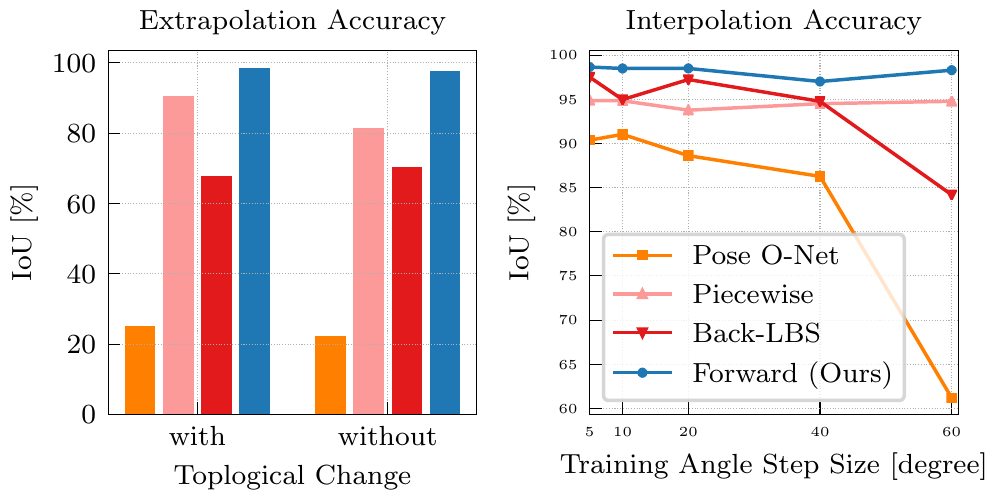}
    \vspace{-2em}
    \caption{\textbf{Quantitative Results on 2D Toy Experiment.} \emph{Left:} For pose extrapolation, our method outperforms all baselines on both test cases, with and without topological changes. \emph{Right:} When interpolating, the performance gap increases as training angles are sampled more sparsely. \vspace{-1em}}
    \label{fig:toy_plot}
\end{figure}

%% file: results/3d_naked/numbers_with_dnerf.tex
\begin{table*}
\centering
\resizebox{\textwidth}{!}{
\begin{tabular}{l|ccccc|ccccc|ccccc|ccccc}
\toprule
{} & \multicolumn{10}{c|}{Within Distribution} & \multicolumn{10}{c}{Out of Distribution} \\
\cline{2-21}
{} & \multicolumn{5}{c|}{IoU bbox} & \multicolumn{5}{c|}{IoU surface} & \multicolumn{5}{c|}{IoU bbox} & \multicolumn{5}{c}{IoU surface}\\
{Subject} & {P.-ONet} & \update{Back-D} & \update{Back-LBS} & {NASA} & {Ours} & {P.-ONet} & \update{Back-D} & \update{Back-LBS} & {NASA} & {Ours} & {P.-ONet} & \update{Back-D} & \update{Back-LBS} & {NASA} & {Ours} & {P.-ONet} & \update{Back-D} & \update{Back-LBS} & {NASA} & {Ours}\\
\midrule
50002    & 84.80\%  & 87.89\% & 47.34\%    & 96.56\%  & \textbf{97.50\%}  & 63.86\%  & 66.42\% & 85.41\%  & 84.02\%  &\textbf{89.57\%}  & 60.61\% & 70.02\% & 73.42\% & 87.71\%  & \textbf{94.51\%}  & 31.94\%  &39.84\% & 71.01\%  & 60.25\%  & \textbf{79.75\%} \\
50004    & 80.09\%  & 84.52\%  & 93.53\%   & 96.31\%  & \textbf{97.84\%}  & 57.79\%  & 59.93\% & 88.07\%  & 85.45\% & \textbf{91.16\%} &  55.44\% &64.63\% & 65.17\%   &  86.01\% &  \textbf{95.61\%} & 34.26\%  & 38.62\% & 69.43\%  &  62.53\% & \textbf{83.34\%} \\
50007    & 88.31\%  & 89.09\%  &  50.13\% & 96.72\%  & \textbf{97.96\%}&  67.14\% & 68.02\% & 83.46\% & 86.28\%  & \textbf{91.02\%} &  40.53\% & 59.68\% & 62.66\% &80.22\% &  \textbf{93.99\%} & 17.80\% & 34.76\%  & 59.53\%  &51.82\% & \textbf{77.08\%}\\
50009    &  71.67\% & 74.75\% &  65.36\%  & 94.94\%  & \textbf{96.68\%} &  50.87\% & 53.96\% & 85.38\%  & 84.52\%  & \textbf{89.19\%} &  38.17\% & 50.18\% &  63.34\%  & 78.15\% & \textbf{91.22\%}  &  23.24\% & 30.85\% &  64.40\%  & 55.86\%  & \textbf{75.84\%} \\
50020    &  69.21\% & 73.37\% &  93.04\% & 95.75\%  & \textbf{96.27\%}  & 48.73\%  & 53.72\% &  86.03\% & 87.57\%  & \textbf{88.81\%} &  42.66\% & 52.43\% &  64.98\%  & 83.06\% & \textbf{93.57\%} & 26.56\%  &  33.62\% &  68.24\%  &62.01\%  &\textbf{81.37\%} \\
50021    &  79.30\% & 79.48\% & \textbf{96.86\%} & 95.92\%  & \textbf{96.86\%} & 57.80\%  & 64.02\% & 89.96\% & 87.01\%  & \textbf{90.16\%} & 45.50\%   &  58.99\% &  69.89\% & 81.80\% & \textbf{93.76\%}  & 29.07\%  & 37.19\% & 61.69\%   & 65.49\% & \textbf{81.49\%} \\
50022    &  86.60\% &90.59\% & 97.60\%  & 97.94\%  &\textbf{97.96\%} & 66.82\% & 74.27\%  & \textbf{93.51\%} & 91.91\%  & 92.06\% &  52.17\% & 60.41\% & 67.83\%  &  87.54\% & \textbf{94.67\%}  &  33.00\%   & 34.71\% & 73.46\% & 70.23\%  & \textbf{83.37\%} \\
50025    &  80.14\% & 79.81\% & 95.28\%  & 95.50\%  &\textbf{97.54\%}& 59.47\%  & 60.37\% &  87.33\% & 86.19\%  & \textbf{91.25\%} & 52.78\%  & 56.93\% & 68.91\%  &  83.14\% & \textbf{94.48\%}  &  31.37\%  & 34.49\% &  70.60\% &  60.88\% & \textbf{82.48\%} \\
50026    &  79.39\% & 84.58\% &  97.32\% & 96.65\%  & \textbf{97.64\%} & 60.52\%  & 64.07\% &  90.17\% & 87.72\%  & \textbf{91.09\%} &  56.09\% & 64.33\% & 65.20\%  &  84.58\% &\textbf{94.13\%} &  32.07\% & 37.71\% &  71.85\% & 59.78\% & \textbf{80.01\%}\\
50027    &  73.91\% & 76.71\% &  80.33\% & 95.53\%  & \textbf{96.80\%} &  53.91\% & 57.46\% &  85.04\% & 86.13\%  & \textbf{89.47\%} & 48.22\% & 57.00\% & 67.86\%  &  83.97\% &  \textbf{93.76\%} &  27.56\% & 32.56\% & 70.55\%  &  61.82\% & \textbf{81.81\%}\\
\bottomrule
Avg.     & 79.34\%  & 82.08\% & 81.68\%  & 96.14\%  &  \textbf{97.31\%} & 58.61\%  & 62.22\% &  87.44\% & 86.98\%  & \textbf{90.38\%} & 49.21\%  & 59.46\% &  66.93\% &  83.16\% & \textbf{93.97\%} & 28.69\% & 35.44\% &  68.93\% &   60.21\%& \textbf{80.65\%}\\
\bottomrule
\end{tabular}
}
\caption{\textbf{Quantitative Results on Minimally Clothed Humans.} \update{The mean IoU of uniformly sampled points in space (IoU bbox) and points near the surface (IoU surface) are reported. Our method outperforms all baselines including NASA~\cite{Deng2020ECCV}. Improvements are more pronounced for points near the surface, and for poses outside the training distribution.}\vspace{-1em}}
\label{tab:3d_naked}
\end{table*}

%% file: results/3d_naked/visuals_with_dnerf.tex
\begin{figure*}
\begin{center}
\setlength\tabcolsep{1pt}
\newcommand{\crop}{0.8cm}
\newcommand{\cropsmall}{0.4cm}
\newcommand{\height}{2.40cm}
\renewcommand{\arraystretch}{0}
\begin{tabularx}{\linewidth}{ l cc| ccccc }
\rotatebox{90} {~~~~~Pose-ONet~~~~~}
                        &   \includegraphics[height=\height, trim={-1cm 0 {\cropsmall} 0},clip]{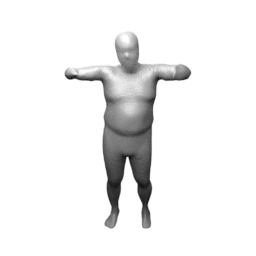}
                        &   \includegraphics[height=\height, trim={{\crop} 0 {\crop} 0},clip]{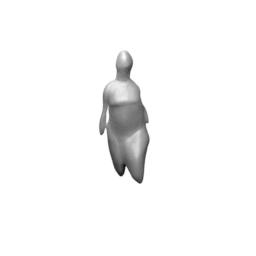}
                        &\includegraphics[height=\height, trim={-1cm 0 {\cropsmall} 0},clip]{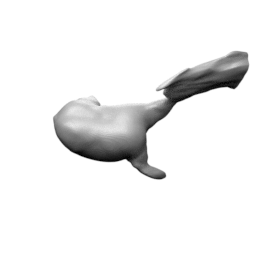}
                        &  \includegraphics[height=\height, trim={{\cropsmall} 0 {\cropsmall} 0},clip]{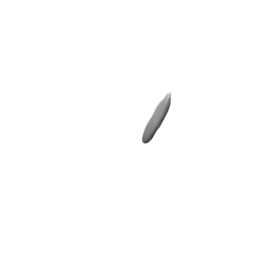}
                        &   \includegraphics[height=\height, trim={{\cropsmall} 0 {\cropsmall} 0},clip]{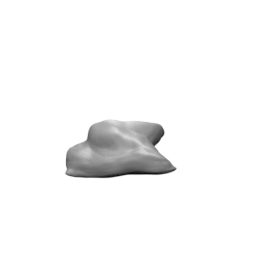}
                        &   
                        &  \includegraphics[height=\height, trim={{\cropsmall} 0 {\cropsmall} 0},clip]{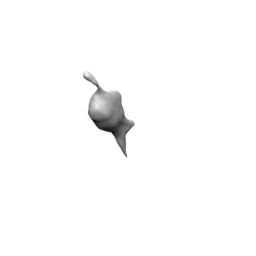}\\
\rotatebox{90} {~~~~~~~\update{Back-D}~~~~~~~} 
                        &   \includegraphics[height=\height, trim={-1cm 0 {\cropsmall} 0},clip]{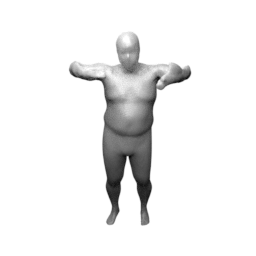}
                        &   \includegraphics[height=\height, trim={{\crop} 0 {\crop} 0},clip]{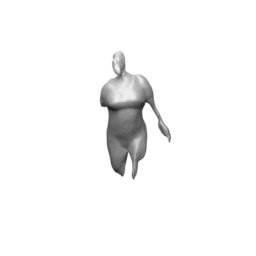}
                        &\includegraphics[height=\height, trim={-1cm 0 {\cropsmall} 0},clip]{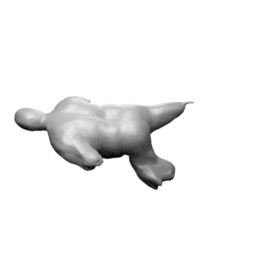}
                        &   \includegraphics[height=\height, trim={{\cropsmall} 0 {\cropsmall} 0},clip]{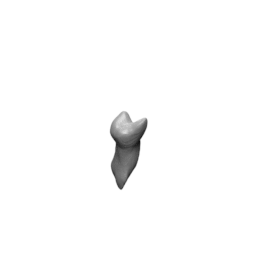}
                        &  \includegraphics[height=\height, trim={{\cropsmall} 0 {\cropsmall} 0},clip]{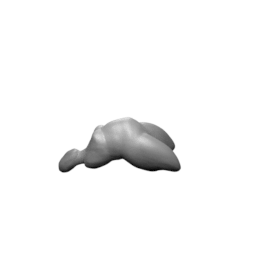}
                        &   \includegraphics[height=\height, trim={{\cropsmall} 0 {\cropsmall} 0},clip]{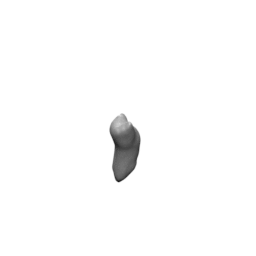}
                        &   \includegraphics[height=\height, trim={{\cropsmall} 0 {\cropsmall} 0},clip]{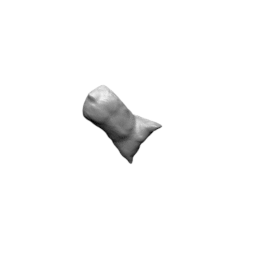}\\
\rotatebox{90} {~~~~~\update{Back-LBS}~~~~~} 
                        &   \includegraphics[height=\height, trim={-1cm 0 {\cropsmall} 0},clip]{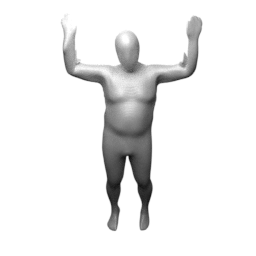}
                        &   \includegraphics[height=\height, trim={{\crop} 0 {\crop} 0},clip]{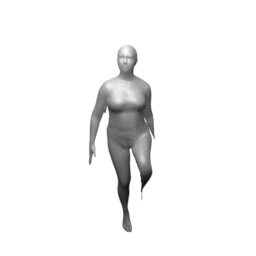}
                        &\includegraphics[height=\height, trim={-1cm 0 {\cropsmall} 0},clip]{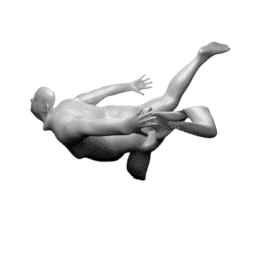}
                        &   \includegraphics[height=\height, trim={{\cropsmall} 0 {\cropsmall} 0},clip]{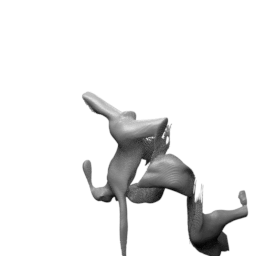}
                        &  \includegraphics[height=\height, trim={{\cropsmall} 0 {\cropsmall} 0},clip]{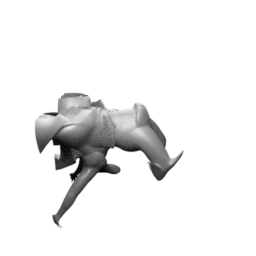}
                        &   \includegraphics[height=\height, trim={{\cropsmall} 0 {\cropsmall} 0},clip]{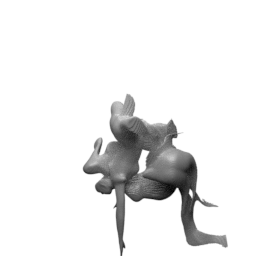}
                        &   \includegraphics[height=\height, trim={{\cropsmall} 0 {\cropsmall} 0},clip]{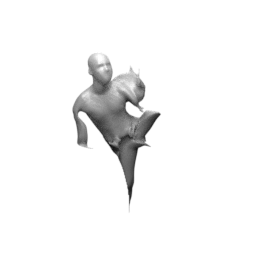}\\
\rotatebox{90} {~~~~~~~NASA~~~~~~}
                        &   \includegraphics[height=\height, trim={-1cm 0 {\cropsmall} 0},clip]{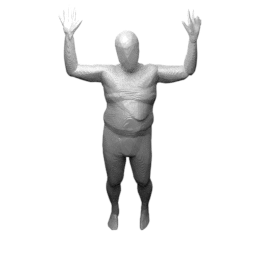}
                        &   \includegraphics[height=\height, trim={{\crop} 0 {\crop} 0},clip]{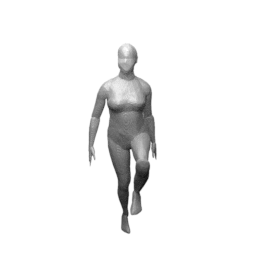}
                        &\includegraphics[height=\height, trim={-1cm 0 {\cropsmall} 0},clip]{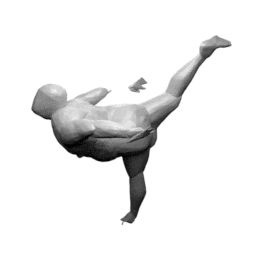}
                        &   \includegraphics[height=\height, trim={{\cropsmall} 0 {\cropsmall} 0},clip]{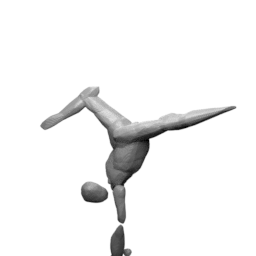}
                        &   \includegraphics[height=\height, trim={{\cropsmall} 0 {\cropsmall} 0},clip]{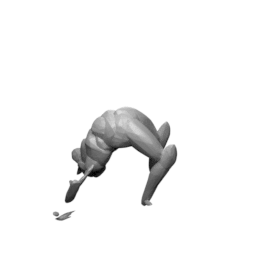}
                        &   \includegraphics[height=\height, trim={{\cropsmall} 0 {\cropsmall} 0},clip]{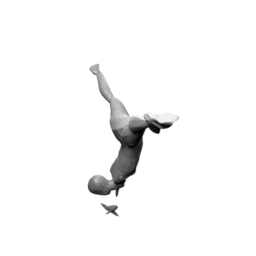}
                        &   \includegraphics[height=\height, trim={{\cropsmall} 0 {\cropsmall} 0},clip]{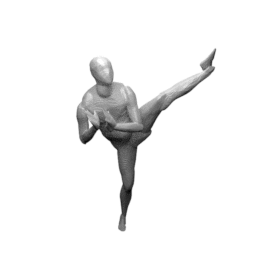}\\
\rotatebox{90} {~~~~~~~~~~Ours~~~~~~~~}
                        &   \includegraphics[height=\height, trim={-1cm 0 {\cropsmall} 0},clip]{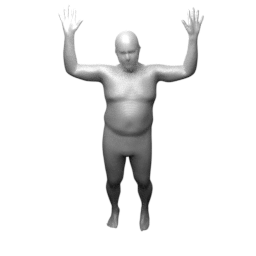}
                        &   \includegraphics[height=\height, trim={{\crop} 0 {\crop} 0},clip]{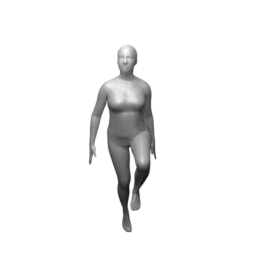}
                        &\includegraphics[height=\height, trim={-1cm 0 {\cropsmall} 0},clip]{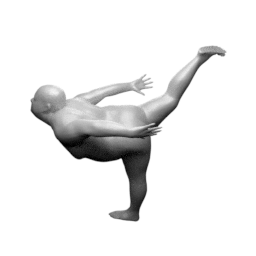}
                        &   \includegraphics[height=\height, trim={{\cropsmall} 0 {\cropsmall} 0},clip]{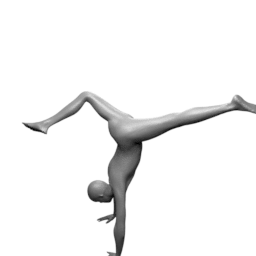}
                        &   \includegraphics[height=\height, trim={{\cropsmall} 0 {\cropsmall} 0},clip]{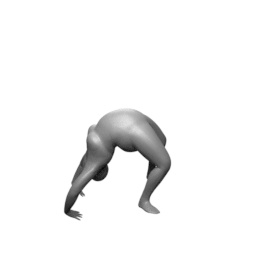}
                        &   \includegraphics[height=\height, trim={{\cropsmall} 0 {\cropsmall} 0},clip]{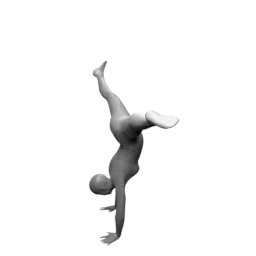}
                        &   \includegraphics[height=\height, trim={{\cropsmall} 0 {\cropsmall} 0},clip]{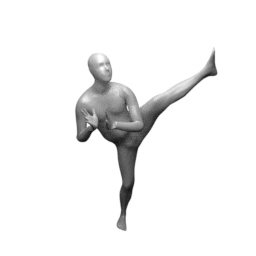}\\
\rotatebox{90} {~~~Ground Truth~~~}  
                        &   \includegraphics[height=\height, trim={-1cm 0 {\cropsmall} 0},clip]{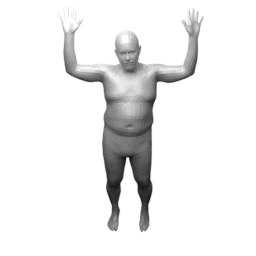}
                        &   \includegraphics[height=\height, trim={{\crop} 0 {\crop} 0},clip]{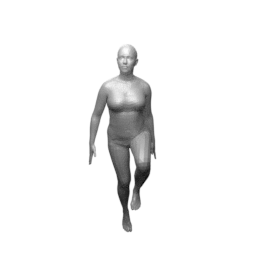}
                        &\includegraphics[height=\height, trim={-1cm 0 {\cropsmall} 0},clip]{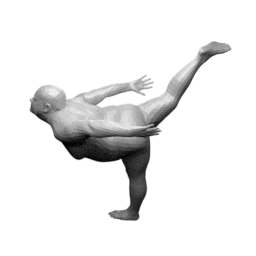}
                        &   \includegraphics[height=\height, trim={{\cropsmall} 0 {\cropsmall} 0},clip]{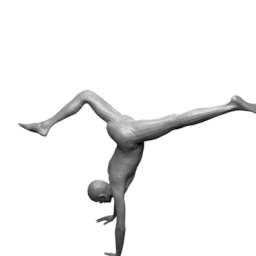}
                        &   \includegraphics[height=\height, trim={{\cropsmall} 0 {\cropsmall} 0},clip]{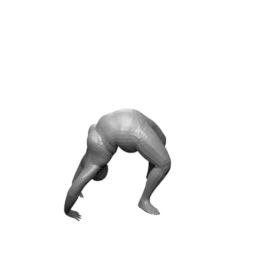}
                        &   \includegraphics[height=\height, trim={{\cropsmall} 0 {\cropsmall} 0},clip]{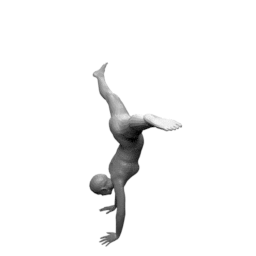}
                        &   \includegraphics[height=\height, trim={{\cropsmall} 0 {\cropsmall} 0},clip]{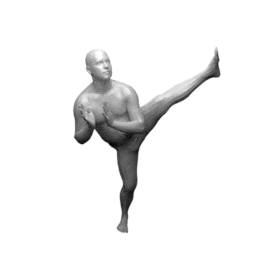}\\
                        &\multicolumn{2}{c}{\centering Within Distribution} &\multicolumn{5}{c}{\centering Out of Distribution}\\

\end{tabularx}
\vspace{-0.5em}
\caption{\textbf{Qualitative Results on Minimally Clothed Humans.} Our method produces results similar to the ground-truth with correct body pose and plausible local details, both for mild poses within the training distribution and more extreme poses. In contrast, the baseline methods suffer from various artifacts including incorrect poses (Pose-ONet), degenerate shapes \update{(Pose-ONet, Back-D, LBS)}, and discontinuities near joints (NASA) which become more severe for unseen poses.
}
\label{fig:3d_naked}
\end{center}
\end{figure*}

%% file: results/3d_clothed/visuals.tex
\begin{figure*}
	\centering
	\begin{tabularx}{\linewidth}{ccccc }
        \includegraphics[height=0.18\linewidth, trim={0 0 0 0},clip]{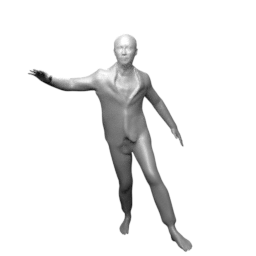}
        &    \includegraphics[height=0.18\linewidth, trim={0 0 0 0},clip]{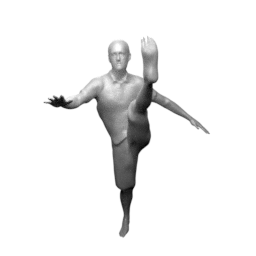}
        &\includegraphics[height=0.18\linewidth, trim={0 0 0 0},clip]{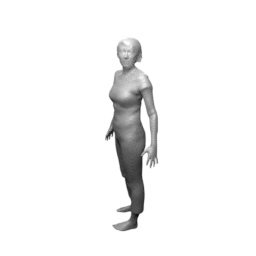}
            &\includegraphics[height=0.18\linewidth, trim={0 0 0 0},clip]{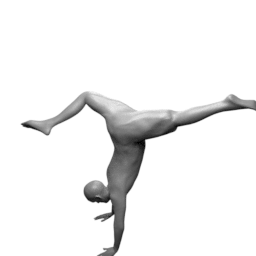}
            
        & \includegraphics[height=0.18\linewidth, trim={1.2cm 0 1.2cm 0},clip]{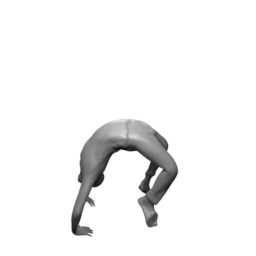}

	\end{tabularx}
	\vspace{-1em}
	\caption{\textbf{Qualitative Results for Clothed Humans.} Our method can model 3D clothed humans in various clothing types, with rich details including wrinkles, \update{and in novel poses.}
	Moreover, our method faithfully learns the non-linear relationship between cloth deformations and body poses.
	On the right, we show a failure case where the cloth does not fall naturally for an extreme, unseen pose. However, note how our method still degrades gracefully in this situation.\vspace{-1em}
	}
\label{fig:3d_clothed}
\end{figure*}

%% file: figures/pose_deformation.tex
\begin{figure}[h!]
    \centering
    \includegraphics[width=\linewidth,trim=0 30 0 50, clip]{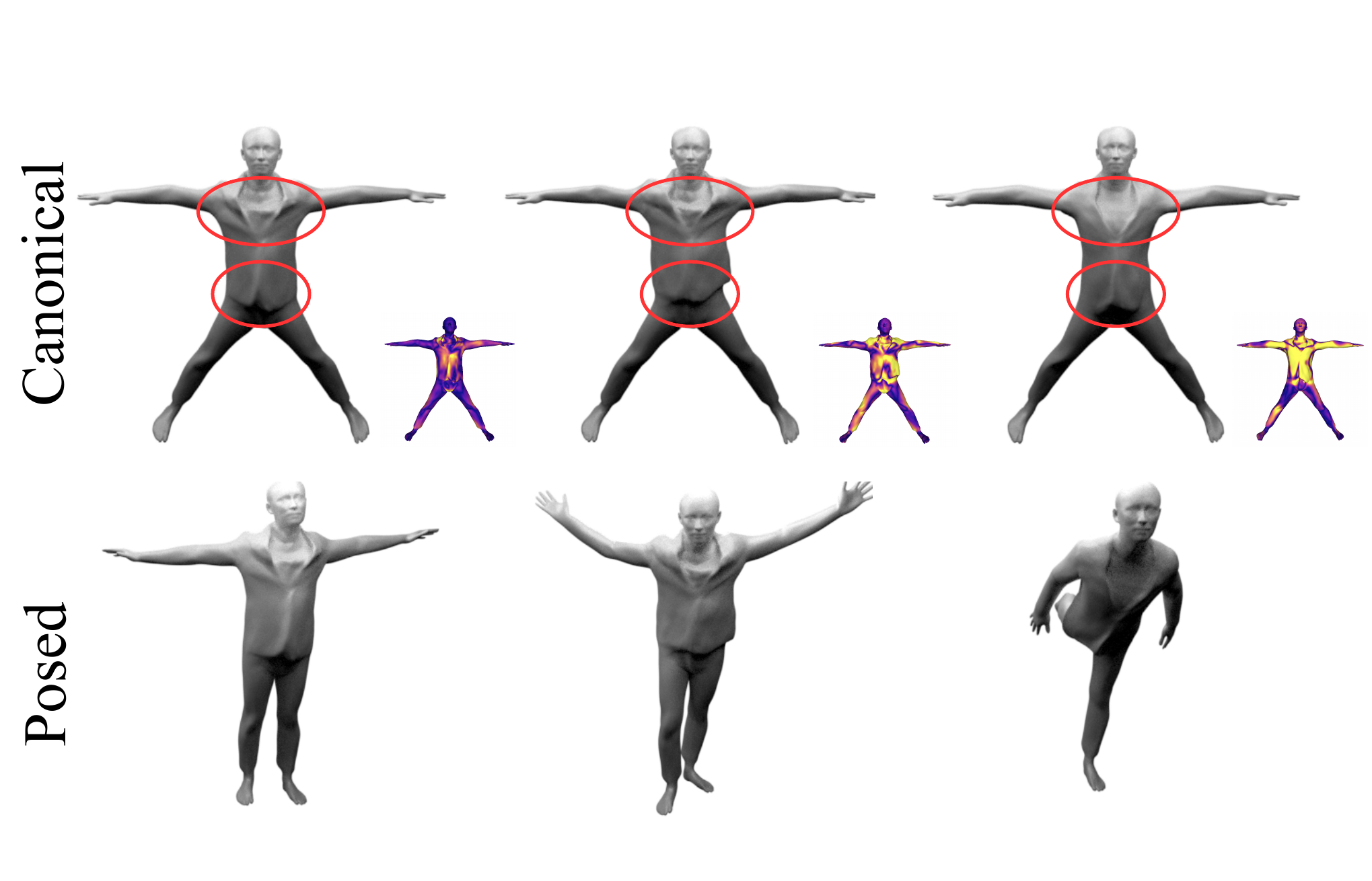}
    \vspace{-1.5em}
    \caption{\update{\textbf{Pose-dependent non-linear deformations (correctives) in canonical space.} The heatmaps show the differences (yellow=large, \change{zoom in for more details}) between the canonical shape for the current pose and the one for the canonical pose, demonstrating the flexibility of the deformations that can be captured by our model.}\vspace{-1.5em}}
    \label{fig:pose_deformation}
\end{figure}

%% file: sections/5_conclusion.tex
\section{Conclusion}
\vspace{-0.5em}
In this paper, we proposed a differentiable forward skinning model for articulating neural implicit surfaces. Our method learns continuous pose-conditioned shapes and skinning weights from  meshes and is able to generate plausible shapes in nearly arbitrary poses. 
We obtain state-of-the-art results on articulated neural implicit representations for 3D human bodies and demonstrate significantly better generalization to unseen poses than the baselines. 
We show \update{SotA} results on challenging cases of (clothed) 3D humans with diverse shapes and poses. 
In future work, we plan to extend our method to learn across subjects and from images only using differentiable rendering \cite{Niemeyer2020CVPR}.

%% file: sections/6_acknowledgement.tex
\update{
{

\boldparagraph{Acknowledgements} Xu Chen and Yufeng Zheng were supported by the Max Planck ETH Center for Learning Systems. Andreas Geiger was supported by the DFG EXC number 2064/1 - project number 390727645.

\vspace{-0.5em}
\boldparagraph{Disclosure} MJB has received research gift funds from Adobe, Intel, Nvidia, Facebook, and Amazon. While MJB is a part-time employee of Amazon, his research was performed solely at, and funded solely by, Max Planck. MJB has financial interests in Amazon, Datagen Technologies, and Meshcapade GmbH.}
}